\documentclass{llncs}

\begin{document}
\title{A Brief State of the Art of CNLs for Ontology Authoring}




\author{Hazem Safwat \and Brian Davis}



\institute{Insight Centre for Data Analytics,\\ National University of Ireland,\\ Galway, Ireland \\ \email{\{hazem.abdelaal, brian.davis\}}@insight-centre.org}


\maketitle

\begin{abstract}
One of the main challenges for building the Semantic web is Ontology Authoring. Controlled Natural Languages CNLs offer a user friendly means for non-experts to author ontologies. This paper provides a snapshot of the state-of-the-art for the core CNLs for ontology authoring and reviews their respective evaluations.
\end{abstract}
\section{Introduction}
 The Semantic Web endeavours to extend the current Web, by enriching information with well defined meaning, which is machine processable \cite{Berners-Lee2001}. This process is heavily dependent on the existence of ontologies, which describe the domain of interest. Formal data representation can be a significant deterrent for non-expert users or small organisations seeking to create ontologies and subsequently benefit from adopting semantic technologies. This challenges researchers to develop user-friendly means for ontology authoring.
 Controlled Natural Languages (CNLs) for knowledge creation and management offer an attractive alternative for non-expert users wishing to develop small to medium sized ontologies. Controlled Natural Languages are defined as ``subsets of natural language whose grammars and dictionaries have been restricted in order to reduce or eliminate both ambiguity and complexity''\cite{Schwitter-Tilbrook}. The goal of this paper is to provide a snapshot overview of the state-of-the-art with respect to CNLs for the Semantic web. However, for a broader review of the CNLs literature in general, we refer the reader to \cite{kuhn2014cl}. In the remainder of this paper, Section 2, provides an overview of the core CNLs players for the Semantic web. In Section 3, generation driven CNLs will be discussed. Section 4, discusses evaluation of different CNLs, and finally Section 5 offers analytic conclusions.
\section{Main CNLs for the Semantic Web}
\subsection{Attempto Controlled English ACE}
A well known approach involving CNL translation into First Order Logic (FOL) is the popular CNL, \textbf{Attempto Controlled English}\footnote{\url{http://www.ifi.unizh.ch/attempto/}, accessed, Thu 25 Jul 2013 16:54:32 IST} (ACE) \cite{fuchs96attempto}. It is a subset of standard English designed for knowledge representation and technical specifications, and constrained to be unambiguously machine readable into discourse representation structures, a form of first-order logic (ACE can also be translated into other formal languages.)  
 ACE is a mature CNL and has been in development since 1995 for over fourteen years \cite{kuhn2010}. It was first introduced by Fuchs and Schwitter \cite{Fuc96a}. Over forty articles have been published by the Attempto group and over 500 articles contain the term ``Attempto Controlled English'' on Google Scholar, \cite{kuhn2010}.  
ACE is a general purpose CNL and is not restricted to any specific domain. The grammar of ACE is perhaps the most expressive in that it can parse a variety of syntactic phenomena in comparison to other CNLs.  ACE caters for instance for relative clauses, coordinated noun phrases, coordinated adverbial and adjectival phrases, numerical and distributed quantifiers, negation, conditional sentences and some anaphoric pronouns\footnote{\url{http://attempto.ifi.uzh.ch/site/docs/ace/6.5/ace\_constructionrules.html},accessed, Thu 25 Jul 2013 16:54:32 IST}.

 ACE Web Ontology Language known as ACE OWL, a sublanguage of ACE, as a means of writing formal, simultaneously human-and-machine-readable summaries of scientific papers \cite{Kaljurand06} \cite{Kuhn06}.
 ACEView is a plugin for the Prot\`{e}g\`{e} editor\footnote{\url{http://Protege.stanford.edu/}, accessed, Thu 25 Jul 2013 16:54:32 IST} \cite{kaljurand:owled2008}. It empowers Prot\`{e}g\`{e} with additional interfaces based on the ACE CNL in order to create, browse and edit an ontology. The user can also query the ontology using ACE questions to access newly asserted facts from the knowledge base. ACE has also served as the basis for other applications such as interface language for a first-order reasoner \cite{fuchs2003reasoning}, a query language for the Semantic Web \cite{bernstein2004talking}, an application for the partial annotation of Webpages \cite{fuchs2007web} and  the usage of ACE for producing summaries within the biomedical domain \cite{kuhn2006improving}. 
 A recent development is the translation of a complete collection of paediatric guideline recommendations into ACE \cite{shiffman2009cnlmain}.
In addition, \textbf{AceWiki} \cite{Kuhn2008} is a monolingual CNL based semantic wiki that takes advantage of ACE for its syntactically user friendly formal language, and of OWL frameworks for applying classification and querying. The AceWiki content is based on ACE predictive editor notation grammar called codeco \cite{Kuhn2012}. The main benefit of codeco is that it can translate all AceWiki content to OWL.
\subsection{Grammatical Framework GF}
Grammatical Framework is an implementation framework for multiple CNLs \cite{AngelovR09} and \cite{ranta04:gf}. GF can cope with a variety of CNLs as well as boost the development of new ones. In \cite{AngelovR09}, the authors reverse engineer ACE for GF in order to demonstrate how portable CNLs are to the GF framework as well as how CNLs can be targeted to other natural languages. ACE is ported from English to five other natural languages.  In short, the core advantage of GF is its multilingualism in that its primary task is domain specific knowledge based Machine Translation (MT) of controlled natural languages. It adds a syntax formalism to the logical framework which defines realisations of formal meanings as concrete linguistic expressions. The semantic model is called the \textit{abstract syntax} while the syntactic realisation functionality is called \textit{concrete syntax}. The authors state that GF is multilingual, in that one abstract syntax, acting as an interlingual, can be (given a concrete syntax for one or more source languages) re-targeted to several languages. The GF libraries now contain a collection of wide coverage grammars for over 15 natural languages. There is an increasing activity with respect to the GF development and a vibrant open source community, which continues to create language resources for GF. The success is also due to the European project, MOLTO (Multilingual On-Line Translation)\footnote{\url{http://www.molto-project.eu/}, accessed, Thu 25 Jul 2013 16:54:32 IST}. This has boosted the uptake of GF and resulted in many comprehensive applications. GF applications range from mathematical proofing, dialog systems, patent translation \cite{moltopatent}, multilingual wikis and multilingual generation in the culture heritage domain \cite{AngelovR09}\cite{Dana200815}. In addition, there have been recent efforts to cater for semantic web ontologies in GF. In \cite{AngelovE10}, the authors develop a conversion tool for compiling axioms in the SUMO ontology \cite{Niles2001} written in the KIF language \cite{genesereth1992} to GF abstract syntax.  In addition, the authors produce CNL from the ontology and allow users to edit SUMO axioms in CNL.  SUMO contains natural language templates for Natural Language Generation (NLG), which were processed and covered into GF concrete syntax. It permits language generation for up to 10 languages, but the templates were lacking  with respect to morphological realisation for languages other than English. GF compensates for these deficits and a fraction of the  English CNL generated was ported to both French and Romanian. Other work in this context involves multilingual generation from a knowledge base within the cultural heritage domain \cite{dannells2012multilingual}. Although GF has no specific CNL, one could argue that its growing open source community may result in GF becoming the de-facto open source general framework for developing resources for engineering multilingual CNLs.
    In \cite{Kuhnkaljurand2013}, the authors introduce a multilingual extension of the previously mentioned AceWiki called \textbf{AceWiki-GF}, where users can get all the benefits of AceWiki in addition to the multilingual environment. The implementation was done by modifying the original AceWiki to include GF multilingual Ace grammar, GF parser, GF source editor, and GF abstract tree set. This study included an evaluation about the accuracy of translation in AceWiki-GF. The evaluation showed that the translation accuracy was acceptable, although some errors due to different reasons in terms of Resource Grammar Library (RGL), where incorrect use of RGL by mixing regular and irregular paradigms, using unnatural phrases to native speakers, and negative determiners. The authors promised a more detailed evaluation in the future work. 
\subsection{Other CNLs}
\textbf{RABBIT Controlled English} is a well known implementation \cite{rabbit08}. It  is essentially an extension of Controlled Language for Ontology Editing CLOnE \cite{Funk07}, but is much more powerful with respect to grammar expressiveness and ontology authoring capabilities. Like CLOnE, Rabbit is implemented using the GATE framework \cite{Cun01b}. Rabbit was developed by the national mapping agency in Great Britain - Ordnance Survey.  Rabbit can be converted to OWL\footnote{\url{http://www.w3.org/TR/owl-features/}, accessed, Thu 25 Jul 2013 16:54:32 IST} to provide natural language support for ontology authoring.  OWL development is not the primary objective of Rabbit. It is primarily a vehicle for capturing, representing and communicating knowledge in a form that is easily understood by domain experts. There are three broad types of sentences in Rabbit - declarations, axioms and import statements. Interestingly, a given class or concept can refer to  a specific ontology in Rabbit i.e. one can refer to the animal \texttt{Duck} within a specific  ontology - \texttt{Waterfowl} as opposed to a default  ontology. Therefore, more than one ontology can be referenced in the Rabbit language \cite{rabbit08}. Rabbit attempts to cater for property restrictions such as transitivity and symmetry, but as the authors themselves argue that such concepts are ``not aligned to the way people think'' and that there is no ideal solution to creating natural language equivalents to property restrictions.  Arguably, these issues should be dealt with by support from the ontology engineer and not the domain expert directly.

    \textbf{Rabbit to OWL Ontology authoring ROO} is an editing tool seeks to cater for the entire ontology engineering process  \cite{dimitrova08}.  It was developed by the University of Leeds and is an open source Java based plug-in for Prot\'{e}g\'{e}. ROO supports the domain expert in creating and editing ontologies using Rabbit. The authors argue that CNL interfaces tend to ignore the ontology construction process. The design of the ROO interface is based on Ordnance Survey proposed ontology development methodology called \textit{Kanga} \cite{kan}. Domain experts are involved in the early stages of the ontology engineering process and engage in the conceptualisation of the ontology, while the ontology engineer is involved at the end stages and focus on the logical level of the ontology. The work of \cite{dimitrova08}, gives a good overview of Rabbit's expressiveness with respect to Rabbit's syntax patterns and their corresponding ontology mappings such as existential quantifiers, union, disjointness and cardinality. A new intelligent model was integrated to ROO to understand the user actions and give feedback accordingly. The model was introduced in \cite{Denaux2012} to resolve the modelling errors, by providing a framework for semantic feedback when adding a new fact to an existing ontology. The new framework extends the syntactic analysis performed by Rabbit through categorizing the new ontological facts into four categories concerning inconsistency and novelty of facts.
  This feedback approach was observed to be repetitive, confusing and sometimes redundant \cite{Denaux2013}. As a result, a new framework with dialogue interfaces was introduced in \cite{Denaux2013} as an extension to Rabbit. It provides more appropriate feedback according to different situations by keeping track of the ontology history. In addition, the inputs of the domain experts are analyzed and an intention is assigned to each input.
\section{Generation driven CNLs}
    \textbf{What you see is what you meant - WYSIWYM}
With respect to ontology driven generation of CNLs or \textit{conceptual authoring}, a well-known implementation which employs the use of NLG to aid the knowledge creation process is \textbf{WYSIWYM} \cite{Power98}. It involves direct knowledge editing with natural language directed feedback. A domain expert can edit a knowledge based reliably by interacting with natural language menu choices and the subsequently generated natural language feedback which can then be extended or re-edited using the menu options. Similar to WYSIWYM, \textbf{GINO} (Guided Input Natural Language Ontology Editor) provides a guided, controlled NLI (natural language interface) for domain-independent ontology editing for the Semantic Web. GINO incrementally parses the input not only to warn the user as soon as possible about errors but also to offer the user (through the GUI) suggested completions of words and sentences---similarly to the ``code assist'' feature of Eclipse \footnote{\url{http://www.eclipse.org/}, accessed, Thu 25 Jul 2013 16:54:32 IST}  with respect to morphological realisation and other development environments \cite{Bernstein06}. 

 \textbf{Round Trip Ontology Authoring ROA}
builds on and extends the existing advantages of the CLOnE software and input language. It generates the entire CNL document first using SimpleNLG that is less sophisticated than WYSIWYM \cite{Gatt2009}. However, it has performed well in user's evaluation \cite{DavisIFTBCH08}.
   
   \textbf{OWL Simplified English} is another WYSIWYM inspired CNL \cite{Power12}.  It is a finite state language for ontology editing. The argument for the finite state approach is that the majority of the OWL expressions created by ontology developers were invariably right branching and hence could be recognised by a finite state grammar. Based on previous studies of ontology corpora, the authors show how the individuals, classes  and properties tend to have distinct Part Of Speech (POS) tags.  Individuals or instances tend to be either proper nouns, common nouns or numbers, while classes are composed mostly of common nouns, adjectives and proper nouns.  Finally, properties tend to open with a verb or auxiliary verb in the present tense.  In paper \cite{Power12}, the authors describe a finite state network that is capable of interpreting the CNL sentences in the grammar with minimal knowledge of content words.
 OWL Simplified English permits the acceptance of  some technical phrases that violate normal English. The language can capture ontology operations such as simple negation, cardinality, object intersection but aims to reduce or eliminate structural ambiguity. We include OWL simplified English as the interface, under construction, is a WYSIWYM based interface.
\section{Evaluation of CNLs}
 With respect to related work, we will review existing CNL research, but in the context of user evaluation. As discussed in Section 2.1, Attempto Controlled English \textbf{ACE} is a well known CNL \cite{fuchs96attempto}. Recently Kuhn \cite{Kuhn2013} described an evaluation framework for CNLs based on Ontographs. Ontographs are a graphical notation to enable tool independent and reliable evaluation of the human understanding of a given knowledge representation language. The author categorises CNLs evaluations into (1) task-based, whereby users are provided with a specific task to complete, and (2) paraphrase-based which are concerned with testing the understandability of the CNL. Ontographs serve as a common basis for testing and comparing the understandability of two different formal languages and facilitate the design of tool-independent and reliable experiments. The author claims that Ontographs are simple and intuitive. They are useful for representing simple logical forms but they do not cater for functions and are restricted to unary and binary predicates. In short, Ontographs serve to test the relative understanding of the core logic for two different formal languages. The experiments compared the syntax of the CNL framework ACE versus OWL framework called simplified Manchester OWL to test which framework is better in terms of, understandability, learning time, and users acceptance. The results showed that users were able to do better classification using ACE with approximately 5\% more accuracy than Manchester OWL, and 4.7 minutes less for learning and testing. Also, in terms of understandability ACE got a higher score than Manchester OWL \cite{Kuhn2013}.
 
  In \cite{EngelbrechtHD09}, the authors undertake a paraphrase-based evaluation to assess whether domain experts without any ontology authoring development can author and understand declaration and axiom sentences in \textbf{Rabbit}. The experiment included 21 participants from the ordnance survey domain and a Rabbit language expert. The participants were given a text that describes a fictional world and were asked to make knowledge statements which were then compared to equivalent statements created by the Rabbit expert. The sentences produced by non-experts were analysed for correctness (with regard to the knowledge captured) by independent experts and were compared to those produced by the Rabbit expert. Interestingly, on average 51\% of the sentences generated at least one error. Furthermore, the most common error was the omission of the quantifier at the beginning of “every” sentence. An evaluation study of \textbf{ROO} was conducted against ACEView \cite{kaljurand:owled2008} where participants from the domains of geography and environmental studies were asked to create ontologies based on hydrology and environmental models, respectively. Both ontology creation tasks were designed to resemble real tasks performed by domain experts at OS. Controls were put in place to eliminate bias and ontologies for both domains, were also produced by the OS to compare against the ROO generated ontologies. The quantitative results were favourable. Although ACEView users were more productive (not in the statistically significant sense), they tended to create more errors in the resulting ontologies. Furthermore, with respect to ROO users, their understanding of ontology modelling improves significantly in comparison to ACEView. Interestingly, but not surprisingly, none of the ontologies produced were usable without post editing. With respect to the extension of ROO in \cite{Denaux2012} the study showed that 91\% of the feedback messages were helpful to the users, and 78\% were informative. However, feedback caused confusion and overwhelming for 10\% of the cases.
 
 An evaluation of \textbf{WYSIWYM} was carried out with 16 researchers and PhD students from the social sciences domain. Users were shown a six minute background video which described the main functionalities of the WYSIWYM interface \cite{HielkemaME08}.  Descriptions of four resources (documents  to associate metadata) were provided to the users. These descriptions were described as paragraphs of English. The goal was to reproduce the descriptions using the WYSIWYM tool. Each subject also received the descriptions in varied order. Four descriptions were given, which were further divided into eight to ten sub tasks. The successful completion of certain sub-tasks was dependent on the preceding  sub-task. Task completion times, number of operations as well as errors including ``avoidable'' errors (which imply the result of an error introduced from a previous sub-task), were measured. The results were encouraging, where users mean completion times  decreased significantly.  Hence, users gained speed over time.  In addition, user feedback was positive, however the results were less positive in comparison to an earlier evaluation of WYSIWYM \cite{Hallett2007}, whereby users completion of tasks was less accurate \cite{HielkemaME08}. Note that the domain ontology was medical as it was in the context for the CLEF\footnote{\url{http://www.clinical-escience.org/}, Retrieved 2008-05-22} project.  Furthermore, the evaluation involved composing SQL queries to a relational database.  More importantly, users from the social sciences field reported that they were overwhelmed by the large number of options available i.e. thirty properties per one object. CLEF was also developed for the well structured domain of medicine while social sciences tends to be more varied with many different theories and approaches. Consequently, the underlying domain ontology can have a large a significant impact on usability.

\section{Conclusion}

With respect to CNLs for ontology authoring we make the following analytic conclusions:
\begin{itemize}
\item  Grammatical Framework, (GF) appears to be gaining momentum in the CNL research community. It is possible that GF, may take on the role of a general architecture for developing controlled languages.  Furthermore, research within the CNL community is turning its attention towards multilingual controlled languages, with recent efforts to generate ACE, using GF, for several European languages.
\item  There has been an increasing tendency towards conducting proper user evaluation for CNLs. While some CNL researchers have conducted task based evaluations, there have been less comparative evaluations across tools.  In general, the CNL community should invest more in conducting strong user evaluations and not to lose track of the end goal - the creation of more  user friendly ontology editing interfaces.
\item A major question is whether a CNL is appropriate for the task? Although, in the context of ontology authoring, CNLs like CLOnE and ACE offer an attractive alternative to ontology editors, we argue that a CNL is not a panacea for formal knowledge engineering. We argue that for these scenarios, there should be a pre-existing use case for a \textit{human orientated} CNL, in other words a restricted vocabulary or syntax for a technical domain either legal, clinical or aeronautics such as ASD Simplified Technical English\footnote{\url{http://www.asd-ste100.org/}, accessed, Thu 25 Jul 2013 16:54:32 IST}. Without such a use case (despite it being possible to adapt a human-orientated CNL to a machine processable CNL), there would be little incentive for users to interact with it. Factors to be taken into account when designing CNLs include, the knowledge creation task complexity, target user (specialist or non expert), the domain (open or specific), available corpora, sample texts, pre-existing language resources or vocabularies, ontologies, multilingualism, requirements for language generation capabilities, and finally, availability of an NLP engineer or computational linguist for development of general purpose CNLs.
\item Other issues include whether to adopt a shallow or deeper NLP approach?  CLOnE and RABBIT \cite{rabbit08} are based on a suite of shallow linguistic analysis tools while Grammatical Framework (GF) and Attempto Controlled English (ACE) are more lexicalised. Furthermore, they are both more powerful with respect to knowledge modelling.  Both GF and ACE are bidirectional, which is extremely useful for surface realisation.  In addition, GF, which is based on the functional language paradigm, can exploit subsumption for free and moreover has an exhaustive bank of application grammars for multiple languages. ACE on the other hand is logic based and has built-in discourse representation structures which are unification based. However, both RABBIT and CLOnE, respectively, as GATE applications, have a number of Semantic Web and Linked Data processing resources available as GATE resources \cite{Cun02b}. In summary, deciding on what CNL or tools to use depends very much on the complexity of both the knowledge creation task and the language modelling task of the CNL as well as the target knowledge representation language and whether there is a need to reuse existing ontologies or vocabularies.  
\item As research into CNLs has been invigorated to a certain degree by the Semantic Web initiative, Semantic Web researchers with an interest in CNLs, should observe lessons learned by previous work in designing CNLs.  Corpus analysis and empirical approaches should be a necessary step when designing a CNL \cite{Grover}.

\end{itemize}
 
\section{Acknowledgements}
This publication has emanated from research conducted with the
financial support of Science Foundation Ireland (SFI) under Grant Number SFI/12/RC/2289
 
\bibliographystyle{splncs}

\end{document}